\documentclass[10pt,twocolumn,letterpaper]{article}

\usepackage{cvpr}              

\usepackage{algorithm}
\usepackage[noend]{algpseudocode}
\usepackage{graphicx}
\usepackage{amsmath}
\usepackage{amssymb}
\usepackage{booktabs}
\usepackage{multirow}

\usepackage[accsupp]{axessibility}  

\usepackage[pagebackref,breaklinks,colorlinks]{hyperref}

\usepackage[capitalize]{cleveref}

\def\cvprPaperID{64} 
\def\confName{CVPR}
\def\confYear{2022}

\DeclareMathOperator*{\mean}{mean}

\begin{document}

\title{\vspace{-1em} Few-Shot Image Classification Benchmarks are Too Far From Reality: \\ Build Back Better with Semantic Task Sampling}

\author{
    \vspace{-22pt} \\
    Etienne Bennequin\textsuperscript{1, 2} \qquad
    Myriam Tami\textsuperscript{1} \qquad
    Antoine Toubhans\textsuperscript{2} \qquad
    C\'eline Hudelot\textsuperscript{1} \\
    \vspace{-8pt} \\
    \textsuperscript{1}CentraleSup\'elec, Universit\'e Paris-Saclay,  France \\ \textsuperscript{2}Sicara, France \\
    \tt\small\{etienneb, antoinet\}@sicara.com \qquad \tt\small firstname.name@centralesupelec.fr 
    
    \vspace{-0.8em}
}



\maketitle

\begin{abstract}
Every day, a new method is published to tackle Few-Shot Image Classification, showing better and better performances on academic benchmarks. Nevertheless, we observe that these current benchmarks do not accurately represent the real industrial use cases that we encountered. In this work, through both qualitative and quantitative studies, we expose that the widely used benchmark \textit{tiered}ImageNet is strongly biased towards tasks composed of very semantically dissimilar classes \eg bathtub, cabbage, pizza, schipperke, and cardoon. This makes \textit{tiered}ImageNet (and similar benchmarks) irrelevant to evaluate the ability of a model to solve real-life use cases usually involving more fine-grained classification. We mitigate this bias using semantic information about the classes of \textit{tiered}ImageNet and generate an improved, balanced benchmark. Going further, we also introduce a new benchmark for Few-Shot Image Classification using the Danish Fungi 2020 dataset. This benchmark proposes a wide variety of evaluation tasks with various fine-graininess. Moreover, this benchmark includes \textit{many-way} tasks (\eg composed of 100 classes), which is a challenging setting yet very common in industrial applications. Our experiments bring out the correlation between the difficulty of a task and the semantic similarity between its classes, as well as a heavy performance drop of state-of-the-art methods on many-way few-shot classification, raising questions about the scaling abilities of these methods. We hope that our work will encourage the community to further question the quality of standard evaluation processes and their relevance to real-life applications.
\end{abstract}

\section{Introduction}

\begin{figure*}[t!]

    \centering
  \begin{subfigure}{0.48\linewidth}
  \centering
    \includegraphics[width=.83\linewidth]{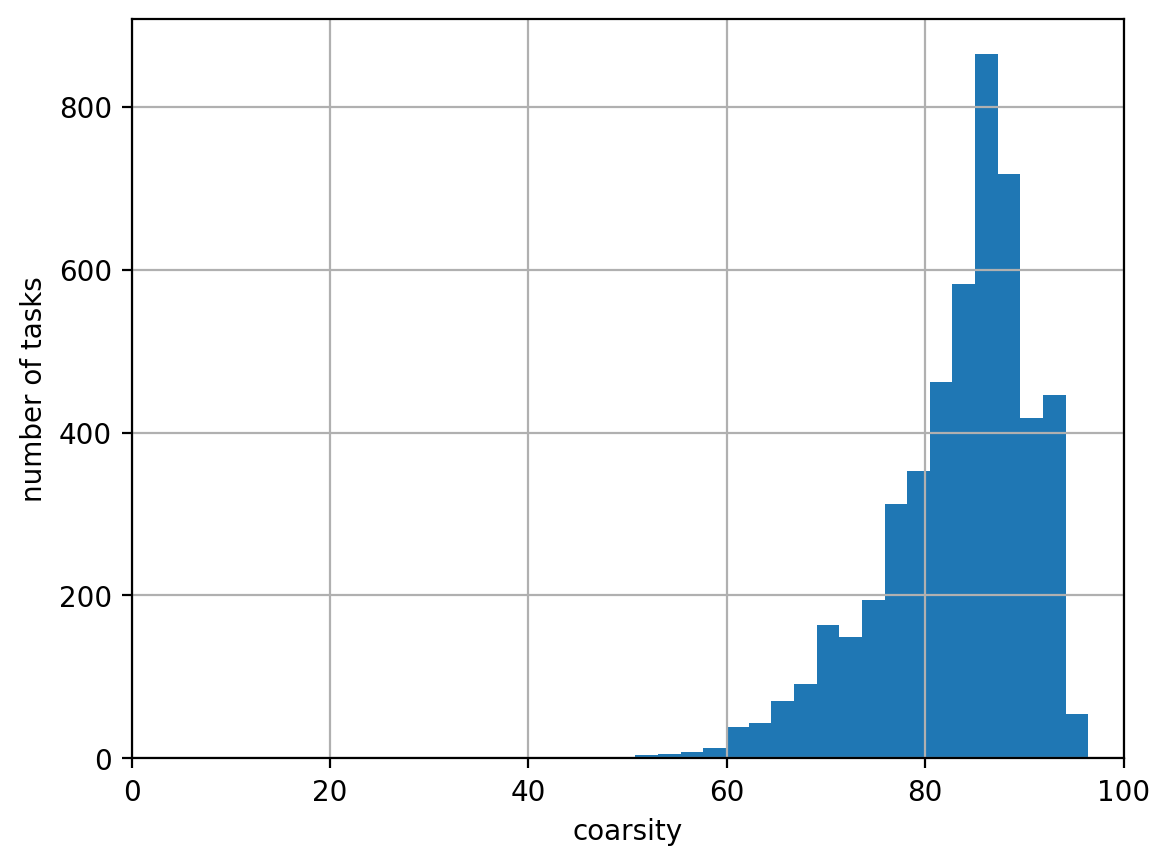}
    \includegraphics[width=.93\linewidth]{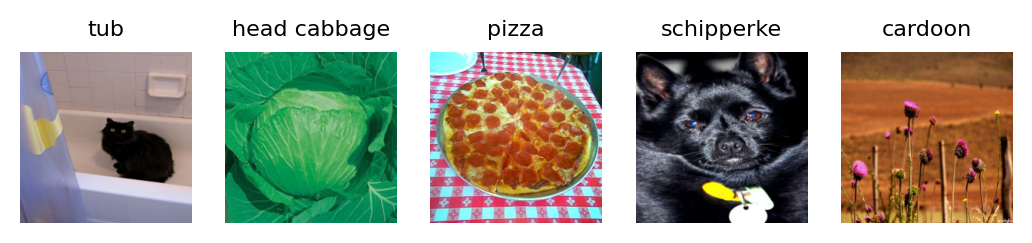}
    \caption{Coarsity histogram (top) and an example of task (bottom) of a testbed designed from \textit{tiered}ImageNet with uniform class sampling. This task presents a coarsity of $85.1$, which is the median coarsity for this testbed. "We really need a machine to distinguish bathroom tubs from cabbage, pizzas, cardoon, and some very specific kind of dog!" said no one in the history of humankind.}
    \label{fig:coarsity-uniform}
  \end{subfigure}
  \hfill
  \begin{subfigure}{0.48\linewidth}
  \centering
    \includegraphics[width=.83\linewidth]{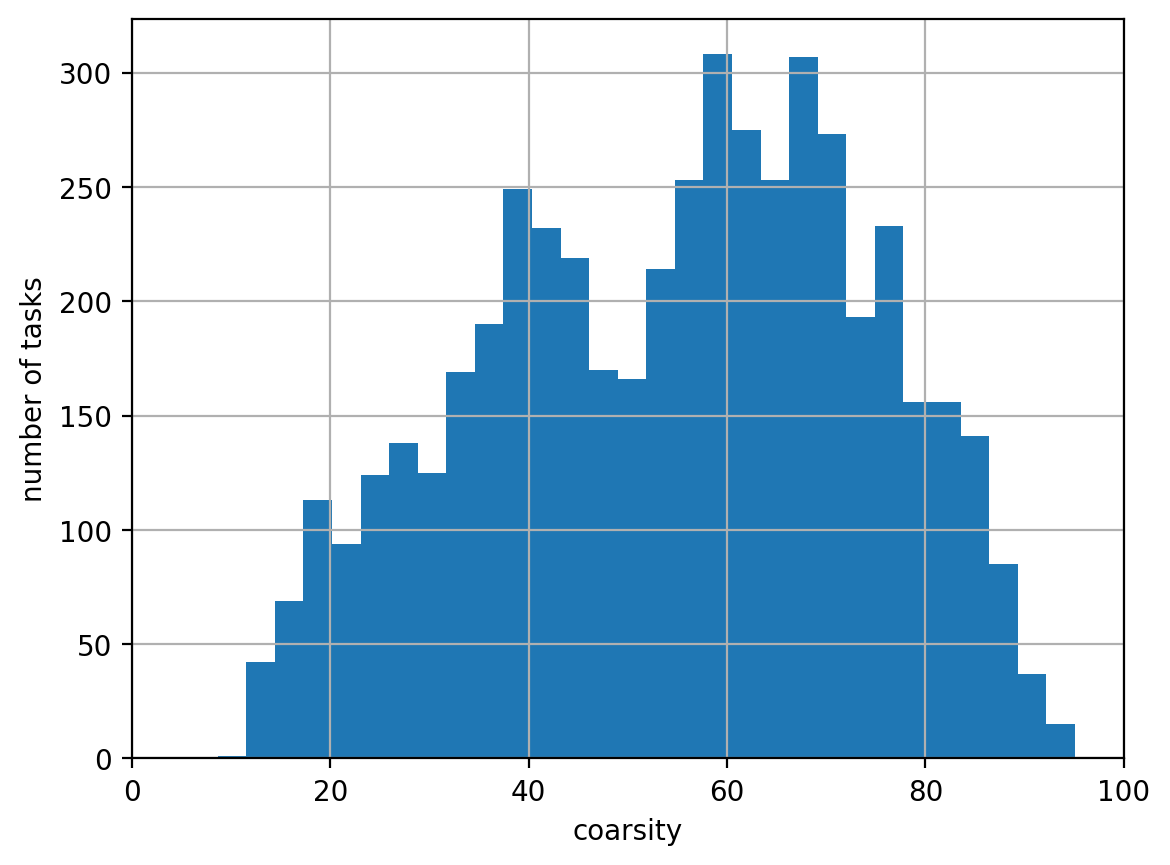}
    \includegraphics[width=.93\linewidth]{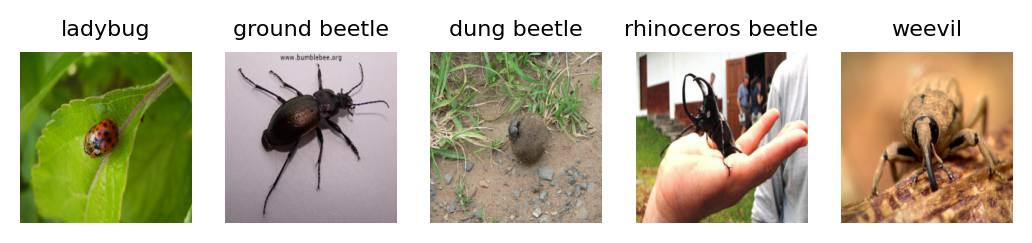}
    \caption{Coarsity histogram (top) and an example of task (bottom) of our testbed \textit{better-tiered}ImageNet. This task presents a coarsity of $15.8$. Tasks with this coarsity never occur in the uniformly sampled testbed, although they are more representative of real few-shot classification use cases.}
    \label{fig:coarsity-better}
  \end{subfigure}
    \caption{Comparison, in terms of our coarsity measure defined in Section \ref{sec:coarsity}, between a testbed designed with uniform class sampling (left) and a testbed designed with semantic awareness (right, ours). Our testbed gives a better representativity to tasks with low coarsity \ie composed of classes semantically relevant to one another.}
    \label{fig:coarsities}
\end{figure*}

Few-Shot Learning is the science of learning new concepts with only a few examples. This task is one of the key abilities of humans but one of the major shortcomings of standard deep learning methods \cite{lake2015human}. It is also required for the success of many industrial applications of computer vision. Businesses that need automatic image recognition do not necessarily possess hundreds of labeled images for each class. It can be because some (or all) classes are rare, or appeared recently, or because classes change every day. In the last three years, our team was involved in a variety of industrial use cases: retrieving a tool in a merchant's catalog, identifying a floor to facilitate recycling, recognizing dishes in a cafeteria given today's menu, or finding the reference of a printed circuit board. In all of these use cases, some or all classes were represented in our database by only 1 to 5 examples. They also had in common that they consisted in recognizing an object among many classes that were semantically similar to one another.

Sadly, because of this, we could not rely on academic benchmarks to identify the most appropriate methods. The first reason is that, as we show in this paper, standard Few-Shot Image Classification benchmarks generate tasks using uniform random sampling from a wide range of semantically dissimilar or unrelated classes (\eg \textit{tiered}ImageNet), which leads to evaluating our models mostly on tasks composed of objects that we would never need to distinguish in real-life use cases (see Figure \ref{fig:coarsity-uniform}). The second reason is that the Few-Shot Learning community has chosen to formalize the Few-Shot Image classification problem as an accumulation of $n$-way $k$-shot classification tasks \ie classifying query images, assuming that they belong to one of $n$ classes for which we have $k$ labeled examples each. In practice, most works compared their methods on benchmarks for which they fixed $n=5$ (sometimes $n=10$) and $k=1$ or $k=5$\footnote{\url{https://paperswithcode.com/task/few-shot-image-classification}}. To the best of our knowledge, only one method was evaluated with $n>50$ \cite{ramalho2019adaptive}. The choice made by the community, while relevant to facilitate experiments in the early stages of Few-Shot Learning research, casts a dark shadow on the robustness of state-of-the-art few-shot learning methods when discriminating between a large number of classes.

How can we improve our current evaluation processes to better fit real-life use cases?
In this paper, we bring out limitations of current Few-Shot Classification benchmarks with both quantitative and qualitative studies and propose new benchmarks to get past these limitations. More specifically:
\begin{enumerate}
    \item We use the WordNet taxonomy \cite{miller1995wordnet} to evaluate \textit{semantic distances} between classes of the popular Few-Shot Classification benchmark \textit{tiered}ImageNet. Based on these semantic distances we put forward the concept of \textit{coarsity} of an image classification task, which quantifies how semantically close are the classes of the task.
    \item We conduct both quantitative and qualitative studies of the tasks generated from the test set of \textit{tiered}ImageNet \ie the tasks composing the benchmark on which most papers evaluate different methods. We show that this benchmark is heavily biased towards tasks composed of semantically unrelated classes.
    \item We harness the semantic distances between classes to generate the improved benchmark \textit{better-tiered}ImageNet reestablishing the balance between fine-grained and coarse tasks. We compare state-of-the-art Few-Shot Classification methods on this new benchmark and bring out the relation between the \textit{coarsity} of a task and its difficulty.
    \item We put forward the Danish Fungi 2020 dataset \cite{picek2022danish} for evaluating Few-Shot Classification models. This dataset offers a wide range of fine-grained classes and therefore allows the sampling of tasks that we deem to be more representative of industrial applications of Few-Shot Learning. We compare state-of-the-art methods on both 5-way and 100-way tasks generated from this dataset. To the best of our knowledge, these are the first published results of few-shot methods on such wide tasks.
\end{enumerate}
All our implementations, datasets and experiments are publicly available\footnote{\url{https://github.com/sicara/semantic-task-sampling}}. We hope that our work will drive the research community towards a better awareness of the biases in few-shot evaluation processes and that the new benchmarks that we propose to counterbalance some of these biases will find echoes in the community and be further improved in future works.

\section{Related Works}

\paragraph{Few-Shot Image Classification} Few-Shot Classification methods historically use episodic training \ie training instances are not images but \textit{episodes} that mimic evaluation tasks that will be sampled from the test set \cite{vinyals2016matching, snell2017prototypical, finn2017model, Ravi16}. Recent works suggest that simple fine-tuning baselines are competitive with sophisticated episodic methods, and focus on developing a good inference method while keeping a classically trained feature extractor \cite{Chen19, dhillon2019baseline, boudiaf2020transductive, liu2020prototype, hu2021leveraging}. Note that some of those methods also consider a \textit{transductive} few-shot classification problem \cite{dhillon2019baseline, boudiaf2020transductive, liu2020prototype, hu2021leveraging} \ie they use unlabeled information from the whole query set to help classification.

\paragraph{Benchmarks} The most simple few-shot classification benchmarks are built respectively on the basic image recognition dataset CIFAR100 \cite{bertinetto2018meta} and Omniglot \cite{lake2011}, which is a dataset of handwritten characters. Rather quickly, \textit{mini}ImageNet \cite{vinyals2016matching} became the reference dataset for Few-Shot Learning. It is a small subset of ImageNet \cite{deng2009imagenet} (60k images in 100 classes, split across classes as train/val/test). Later, \cite{ren2018meta} introduced the larger dataset \textit{tiered}ImageNet, also built from ImageNet but with 608 classes, which are split in a way that preserves the super-category structure of the classes. In this work we focus on \textit{tiered}ImageNet because \textit{mini}ImageNet offers only 20 test classes which clearly do not allow the sampling of any fine-grained 5-way task\footnote{The classes are nematode, king crab, golden retriever, malamute, dalmatian, African hunting dog, lion, ant, black-footed ferret, bookshop, crate, cuirass, electric guitar, hourglass, mixing bowl, school bus, scoreboard, theater curtain, vase, and trifle.}. Note that the classification benchmark CU-Birds 200 \cite{WelinderEtal2010}, which happens to be a fine-grained benchmark, is now also used for Few-Shot Image Classification, especially to study cross-domain robustness \cite{Chen19}.
To our knowledge, all benchmarks built on these datasets follow the same process to generate few-shot classification tasks \ie we sample 5 classes \footnote{Rarely 10, or 20 for Omniglot} uniformly at random from the test set, then sample a fixed number of images per class for the support set and for the query set. 
However, more recently, \cite{triantafillou2019meta} merged 10 computer vision datasets to build a gigantic benchmark for few-shot classification methods: Meta-Dataset. They introduced some randomness in the shape of the tasks (number of ways, shots and queries) and proposed to study the hierarchy of the methods depending on the number of ways and shots. Despite its ability to benchmark methods on incredibly diverse datasets and tasks, Meta-Dataset remains underused by the community compared to a simpler and more lightweight benchmark like \textit{tiered}ImageNet\footnote{21 papers used Meta-Dataset between 2020 and 2021 \textit{v.s.} 92 for \textit{tiered}ImageNet. Source: \textit{PapersWithCode}}. In this work we mitigate the biases of current benchmarks while avoiding additional engineering challenges that would make it harder to adopt our novel benchmarks.

\paragraph{Fine-graininess in Few-Shot Learning} Recent works proposed specific methods for Fine-Grained Few-Shot Image Classification \cite{ruan2021few, tang2020revisiting, xu2021variational, zhu2020multi}. These methods are typically compared on CU-Birds \cite{WelinderEtal2010} or FGVC Aircraft \cite{maji2013fine}. These datasets propose respectively 50 and 25 test classes. In this work, we propose to use Danish Fungi 2020 \cite{picek2022danish}, a fine-grained image classification dataset equipped with 1604 classes, which allowed us to compare methods on tasks composed of a large number of classes. Also note that among many other contributions, the original paper for Meta-Dataset \cite{triantafillou2019meta} proposed a small study of the performance of state-of-the-art few-shot classifiers depending on a measure of \textit{task fine-graininess} on ImageNet. However, they failed to highlight any correlation between the fine-graininess and difficulty of a task, leaving it for future works. We claim to be this future work: compared to \cite{triantafillou2019meta}, we use more precise tools to define the fine-graininess of a task and decorrelate the fine-graininess from the shape of the task (\ie the number of ways). Thereby we successfully show the correlation between fine-graininess and difficulty on \textit{tiered}ImageNet (see Section \ref{sec:results}).

\paragraph{Characterizing classification tasks} Some recent works try and find a way to represent classification tasks so that they can be compared with one another~\cite{nguyen2021similarity, achille2019task2vec}. Here we focus on characterizing tasks using class semantics. \cite{deselaers2011visual} show that on ImageNet, visual and semantic similarities between classes are linked. They measure the semantic similarity with the Jiang \& Conrath pseudo-distance~\cite{jiang1997semantic}, which depends on the WordNet Directed Acyclic Graph and on the number of images in each class. Other methods to evaluate the similarity between categories can be found in \cite{alves2020selection}.

\section{Problem formalization}
\label{sec:problem}

\paragraph{Few-Shot Image Classification}Consider a training set $\mathcal D_{train} = \lbrace (x,y) ~|~ x \in \chi, ~y \in \mathcal{C}_{train} \rbrace$ and a test set $\mathcal D_{test} = \lbrace (x,y) ~|~ x \in \chi, ~y \in \mathcal{C}_{test} \rbrace$ where $\mathcal C_{train}$ and $\mathcal C_{test}$ are their respective class sets, with $\mathcal C_{train} \cap \mathcal C_{test} = \emptyset$, and $\chi$ is the image space. Few-Shot Classification consists in learning on $\mathcal D_{train}$ a classification model that can generalize to the unseen classes in $\mathcal C_{test}$ with only a few training examples per class. 

Given a subset $\mathbf C \subset \mathcal C_{test}$, we note $\mathcal D_{test|\mathbf C} = \lbrace (x,y) \in \mathcal D_{test} ~|~ y \in \mathbf{C} \rbrace$. Then a few-shot classification task $\mathbf T$ is defined by a subset of classes  $\mathbf C \subset \mathcal C_{test}$, a small support set $\mathbf S \subset \mathcal D_{test|\mathbf C}$ of labeled examples (with at least one example for each class \ie $\forall c \in \mathbf C, ~\mathbf S \cap D_{test|\lbrace c \rbrace} \neq \emptyset$) and a query set $\mathbf Q \subset \mathcal D_{test|\mathbf C}$, such that $\mathbf S \cap \mathbf Q = \emptyset$ \ie images in the query set cannot appear in the support set. The task's objective is to maximize the model's accuracy on $\mathbf Q$.

\paragraph{Model evaluation} We define $\mathsf E_{test}(n)$ (and similarly $\mathsf E_{train}(n)$) the set of $n$-way classification tasks that can be sampled from $\mathcal D_{test}$ \ie
\begin{align*}
    \mathsf E_{test}(n) = \lbrace &\mathbf T = (\mathbf S, \mathbf Q) \;| \\
    & \mathbf S, \mathbf Q \subset \mathcal D_{test|\mathbf C}, \\
    & \mathbf S \cap \mathbf Q = \emptyset, \\
    & \mathbf C \subset \mathcal C_{test} \text{ and } |\mathbf C| = n \rbrace
\end{align*}
In practice, most benchmarks are limited to $\mathsf E_{test}(5)$ (which we will note $\mathsf E_{test}$ when there is no ambiguity), and further limited to tasks with 1 or 5 support images per class, and 10 query images per class. The number of possible tasks is still untractable (${\sim}10^{173}$ on \textit{tiered}ImageNet). Therefore we evaluate few-shot classification models on a subset $\Tilde{\mathsf E}_{test} = \lbrace \mathbf T \in \mathsf E_{test} | \mathbf T \sim \mathsf U \rbrace$ with $\mathsf U$ the uniform distribution. This design choice gives the same weight to all tasks from $\mathsf E_{test}$, regardless of how informative they are on a model's ability to perform on real-life few-shot learning problems. In this work, we study alternative distributions for $\Tilde{\mathsf E}_{test}$.



\paragraph{Class sampling and instance sampling} Here we limit ourselves to the sampling of the classes that constitute each episode. We iterate only on the probability distribution over classes that constitute a task. Once the classes are sampled, we use uniform sampling over all instances for each class to constitute the support and query sets.


\section{Building the \textit{better-tiered}Imagenet benchmark}

\subsection{Measuring task coarsity with WordNet taxonomy}
\label{sec:coarsity}

\begin{figure}
    \centering
    \includegraphics[width=.95\linewidth]{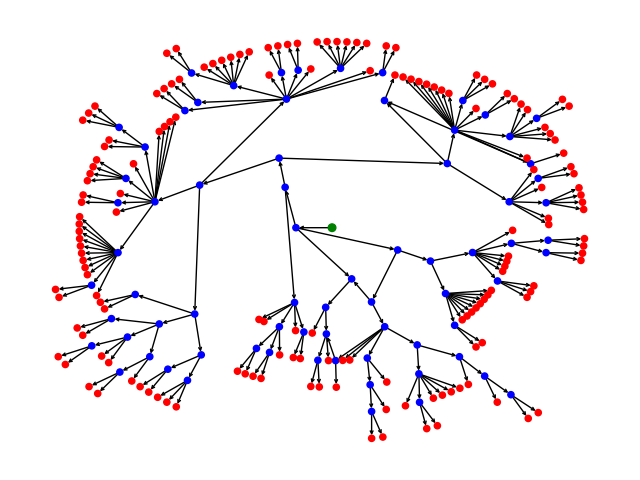}
    \caption{Subgraph of WordNet spanning the 160 classes of \textit{tiered}ImageNet's test set, which are shown in red. The root (in green) corresponds to the concept of ``entity". This is a Directed Acyclic Graph. Best viewed in color.}
    \label{fig:tiered-graph}
\end{figure}

Since \textit{tiered}ImageNet is a subset of ImageNet, its classes are the leaves of a directed acyclic graph which is a subgraph of the WordNet graph \cite{miller1995wordnet} (see Figure \ref{fig:tiered-graph}). Using this graph, it is possible to establish a semantic similarity between classes. We use the Jiang \& Conrath pseudo-distance between classes \cite{jiang1997semantic}, which is defined for two classes $c_1$ and $c_2$ as:
\begin{equation}
    D^{JC}(c_1, c_2) = 2 \log |lso(c_1,c_2)| - (\log |c_1| + \log |c_2|)
\end{equation}
where $|c|$ is the number of instances of the dataset with class $c$, and $lso(c_1, c_2)$ is the lowest superordinate, \ie the most specific common ancestor of $c_1$ and $c_2$ in the directed acyclic graph. Our choice of pseudo-distance\footnote{We call it a pseudo-distance since it is positive, symmetric and separated. However, due to the weak assumption on the directed acyclic graph, it is possible to find $c_1, c_2, c_3$ such that $D^{JC}(c_1, c_3) > D^{JC}(c_1, c_2) + D^{JC}(c_2, c_3)$.} was motivated by the results of \cite{deselaers2011visual}, who showed that this semantic pseudo-distance between classes is correlated with visual similarity between images of these classes on ImageNet. We insist, however, that the definition of the coarsity and the subsequent sampling strategy are agnostic of the choice of the distance, and that other semantic distances may be used in future works.


From this pseudo-distance, we define the \textit{coarsity} $\kappa$ of a task $\mathbf T_{\mathbf C}$ constituted of instances from a set of classes $\mathbf C$ as the mean square distance between the classes of $\mathbf C$ \ie

\begin{equation}
    \kappa(\mathbf T_{\mathbf C}) = \mean_{c_i, c_j \in \mathbf C \\
    c_i \ne c_j} D^{JC}(c_i, c_j)^2
\end{equation}

This coarsity is an indicator of how semantically close are the classes that constitute a task. As shown in \cite{deselaers2011visual}, on datasets derived from ImageNet, this measure is closely linked to the visual similarity between items of these classes.

\subsection{Generate a more informative benchmark using class semantics}
\label{semantic-task-sampling}

As discussed in Section \ref{sec:problem}, finding an appropriate subset of $\mathsf E_{test}$ is a key point to ensure that we provide an accurate evaluation of a model. In the literature, testing tasks are sampled uniformly at random from $\mathsf E_{test}$\cite{vinyals2016matching}. However, we observed that the resulting testbeds are biased towards tasks with high coarsity \ie composed of classes semantically far from each other with respect to the Jiang \& Conrath pseudo-distance (see Figure \ref{fig:coarsity-uniform}). We argue that in practice, few-shot learning models are often used to distinguish between similar objects rather than distinguishing between objects that have nothing to do with one another (\eg circuit boards from circuit boards, carpets from carpets, or people from people). A testbed presenting this type of bias is therefore irrelevant to evaluate a model's ability to solve this family of problems.

In this work, we define a unique, reproducible set of testing tasks to evaluate all models. This testbed is built with a dual objective:
\begin{itemize}
    \item We want tasks with a smooth repartition in terms of coarsity to ensure that the testbed also evaluates the ability of a model to distinguish between classes close to each other. Providing a good span of coarsities also allows to compare models on different types of tasks: a model might be better for coarse tasks but not for fine-grained tasks.
    \item This first objective inherently creates a bias towards classes with many neighbouring classes. However we want our testbed to be balanced, \ie all images must be sampled roughly as many times as the others \footnote{In the case of \textit{tiered}ImageNet, which presents as many images for each class, this is equivalent to ensuring the balance between classes.}.
\end{itemize}

To achieve these goals, we define a semantic task sampler based on the Jiang \& Conrath pseudo-distance. We build an initial potential matrix \cite{liu2020adaptive} $\mathcal{P}_0$ such that $\mathcal{P}_0(i,j) = e^{-\alpha D^{JC}(c_i, c_j)}$ with $\alpha \in \mathbb R_+$ an arbitrary scalar. For the first task, the probability for a pair of classes $(c_i, c_j)$ to be sampled together is proportional to $\mathcal{P}_0(i,j)$.
To enforce that the testbed is balanced, once the $t-1^{\text{th}}$ task is sampled we update the number $\textit{occ}_t(i)$ of occurrences of class $c_i$ in previous tasks. Then we update the potential matrix to penalize classes with higher values of $\textit{occ}_t$:
\begin{equation}
    \mathcal{P}_t(i,j) = \mathcal{P}_0(i,j) \times \exp(-\beta \frac{\textit{occ}_t(i) + \textit{occ}_t(j)}{\max_k(\textit{occ}_t(k))})
\end{equation}
with $\beta \in \mathbb R_+$ an arbitrary scalar. Intuitively, a larger $\alpha$ gives more weight to pairs of semantically close classes, while a larger $\beta$ forces a stricter balance between classes. The class sampling process is detailed in Algorithm \ref{sampling-from-potential}\footnote{We use the convention $\mathcal{P}(i) = \textbf{[}\mathcal P(i,j)\textbf{]}_{j=1}^{|\mathcal C_{test}|}$. Also, $\textbf{occ}$ represents the whole vector, while $\textbf{occ}(i)$ represents the element at index $i$.}.

\begin{figure}
    \centering
    \includegraphics[width=.95\linewidth]{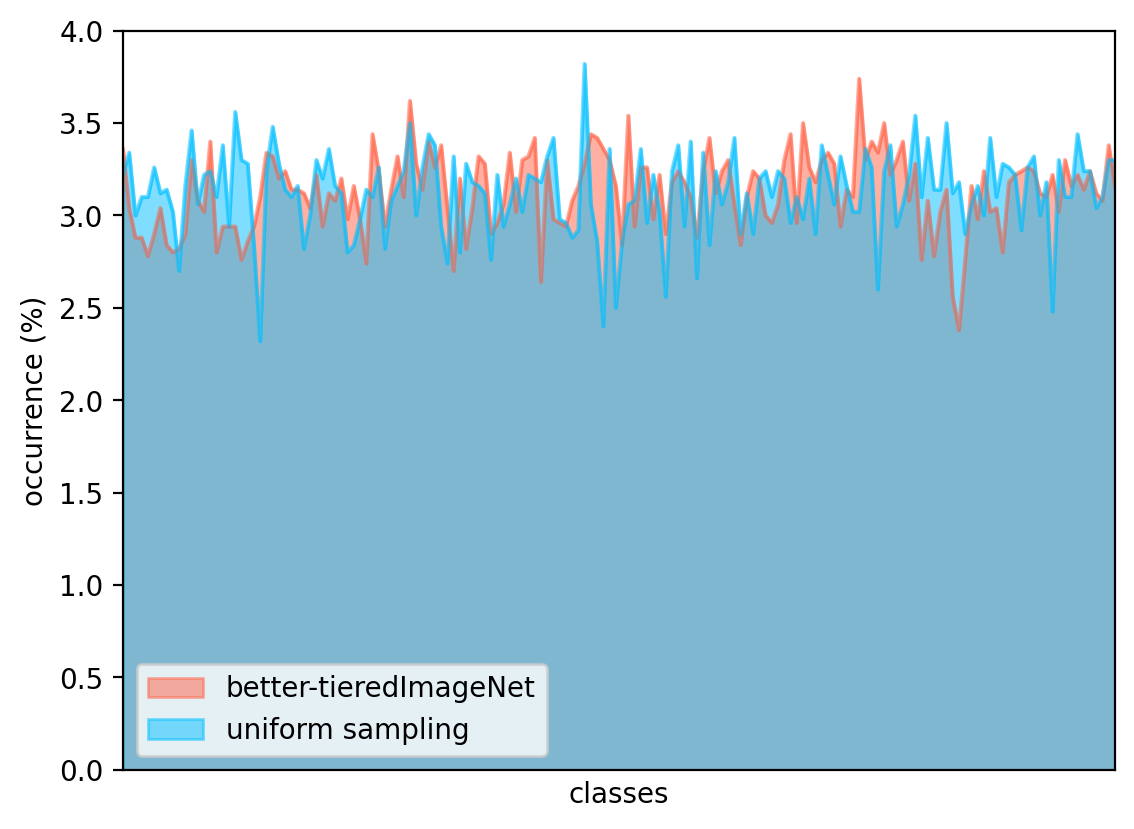}
    \caption{Comparison of class imbalance in testbeds. We show the proportion of tasks containing each one of the 160 classes of \textit{tiered}ImageNet's test set, with uniform task sampling (blue) and our semantic task sampling (red). A perfectly balanced testbed will show a flat line \ie each class is equally represented in the testbed. We see that our sampling does not raise class imbalance compared to uniform sampling.}
    \label{fig:occurrences}
\end{figure}

We then sample instances from these classes uniformly at random. As shown in Figure \ref{fig:coarsity-better}, our 5000-tasks testbed gives far greater representation to fine-grained tasks compared to a uniformly sampled testbed. Our testbed offers a wide range and balance of task coarsities, allowing to test models on both coarse and fine-grained tasks, while the uniformly sampled testbed only allows the evaluation on coarse tasks. Figure \ref{fig:occurrences} shows that the occurrences-based penalty successfully enforces the balance between classes in our testbed.

\begin{algorithm}
\caption{How classes of a task are sampled using the potential matrix, following \cite{liu2020adaptive}}
 \textbf{Input:} potential matrix $\mathcal{P}_0$, number of tasks $T$, number of classes per task $n$ \\
 \textbf{Output:} set of sampled tasks $\mathcal{T}$
\begin{algorithmic}[1] 
    \State $\mathcal{T} \longleftarrow \lbrace \rbrace $
    \State $\mathcal{P} \longleftarrow \mathcal{P}_0$
    \State $\textbf{occ}\longleftarrow [1]_i$
    \For{$t<T$}
        \State $\mathbf p \longleftarrow \exp(-\beta \frac{\textbf{occ}}{\max(\textbf{occ})})$ 
        \State $\mathbf{C} \longleftarrow \lbrace c_0 \rbrace$ with $c_0$ sampled according to a distribution of probability proportional to $\mathbf p$
        \State $\mathbf p \longleftarrow \mathbf p \odot \mathcal{P}_0(c_0) $ 
        \While{$|\mathbf{C}| < n$}
            \State $\mathbf{C} \longleftarrow \mathbf{C} \cup \lbrace c \rbrace$ with $c$ sampled according to a distribution of probability proportional to $\mathbf p$
            \State $\mathbf p \longleftarrow \mathbf p \odot \mathcal{P}_0(c)$
        \EndWhile
        \State $\mathcal{T} \longleftarrow \mathcal{T} \cup \lbrace \mathbf C \rbrace $
        \State $\forall i \in \mathbf C, \textbf{occ}(i) \longleftarrow \textbf{occ}(i) + 1$
    \EndFor
\end{algorithmic}
\label{sampling-from-potential}
\end{algorithm}

\section{Fungi: a large fine-grained dataset for Few-Shot Image Classification}

\subsection{Danish Fungi 2020}

\begin{figure}
    \centering
    \includegraphics[width=\linewidth]{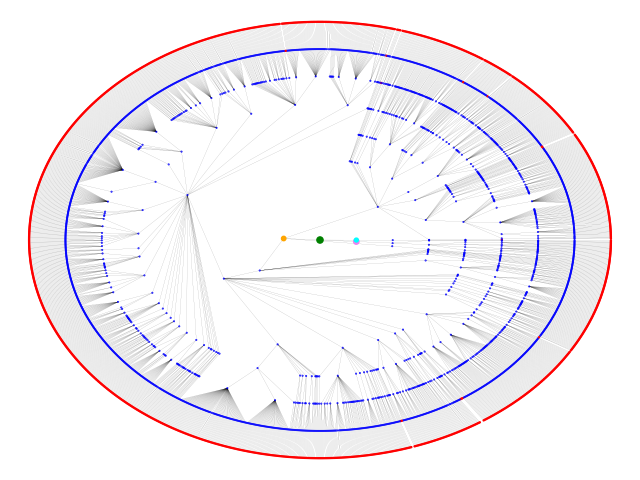}
    \caption{Semantic tree spanning the 1604 classes of DF20 (in red). The root is shown in green. DF20 contains images of species that belong to three kingdoms: Chromista (purple), Protozoa (cyan), and Fungi (orange). As we can see from the tree, the vast majority of classes are spanned from the Fungi kingdom. Best viewed in color.}
    \label{fig:fungi-tree}
\end{figure}

Danish Fungi 2020 (DF20) \cite{picek2022danish} is an image recognition dataset of 295 938 images of fungi distributed in 1604 fine-grained classes, with no overlap with ImageNet. The dataset offers insightful metadata, such as the object's geographical location, habitat, and substrate. DF20's classes are equipped with a seven-level hierarchical structure. Note that Meta-Dataset \cite{triantafillou2019meta} also include a Fungi dataset, from the FGVCx 2018 Fungi classification challenge\footnote{\url{https://github.com/visipedia/fgvcx_fungi_comp}}. This Fungi dataset came from the same source as DF20 but offered fewer images, fewer classes, no metadata, and no taxonomy, which convinced us to push forward DF20. The semantic tree of DF20 is shown in Figure \ref{fig:fungi-tree}.

\subsection{DF20 for Few-Shot Image Classification Benchmark}

\paragraph{Why do we use it?} Following our observations on the high coarsity of tasks sampled from \textit{tiered}ImageNet, we propose to use DF20 as a test set for few-shot image classification models. This dataset allows sampling a wide variety of fine-grained few-shot classification tasks, which are to our experience more representative of real-world applications. It also offers a taxonomy allowing to further study the performance of few-shot models depending on the coarsity of the task. Since we consider the whole dataset as a test set, we allow the comparison of methods with parameters optimized on various training sets, provided that they do not overlap with DF20. This brings the few-shot learning methodology closer to the neighboring field of transfer learning, in which training data is not part of the benchmark \cite{dumoulin2021comparing}. This can also be considered as a generalization of the cross-domain few-shot learning setting, in which we train a model on the training set of one benchmark (\eg \textit{mini}ImageNet) and test it on the testing set of another benchmark (\eg CUB\cite{Chen19}).
Note that in this work, compared methods all share the same backbone with parameters classically trained on ImageNet's training set. This was a convenient baseline, but we insist that this does not make it mandatory for future work to train on ImageNet. In fact, we believe that it is of prior importance to study the effect of the choice of the training data on the few-shot classification model's performance.

\paragraph{How do we use it?}
We sampled four benchmarks: 5-way 1-shot, 5-way 5-shot, 100-way 1-shot, and 100-way 5-shot. The 5-way settings are very common in the few-shot learning literature \cite{triantafillou2019meta}. However, to the best of our knowledge, very few works evaluate their method on \textit{wide} (\ie more than 10-way) few-shot classification tasks\footnote{\url{https://paperswithcode.com/task/few-shot-image-classification}}. We believe that such tasks are at least equally interesting as 5-way tasks. We assume that this setting was avoided by early works because GPU memory limitations made it very hard to use back-propagation on batches mimicking 100-way tasks during episodic training. We claim that these constraints do not justify overlooking such an interesting problem. Specifically for DF20, the task of recognizing an image among a wide variety of fungi makes way more sense than recognizing an image among 5 random species\footnote{As of 2020, experts have identified $\sim$148 000 species of fungi \cite{cheek2020new}.}.

\begin{table*}[t!]

\centering
\begin{tabular}{c|ccccc|c|}
\cline{2-7}
                                              & \multicolumn{6}{c|}{1-shot}   \\ \cline{2-7}
                                              & \multicolumn{5}{c|}{\textit{better-tiered}ImageNet}& \multirow{2}{*}{Uniform testbed}                 \\ \cline{2-6}
                                              & Whole testbed  & 1st. Qtl            & 2nd. Qtl           & 3rd Qtl            & 4th Qtl            &                 \\ \hline
\multicolumn{1}{|l|}{ProtoNet\cite{snell2017prototypical}}                & 53.10 $\pm$ 0.40 & 42.01 & 51.11 & 56.35 & 62.88 & 65.24 $\pm$ 0.35 \\
\multicolumn{1}{|l|}{Finetune\cite{Chen19}}                & 56.60 $\pm$ 0.41 & 44.56 & 54.29 & 60.12 & 67.43 & 69.96 $\pm$ 0.35 \\ \hline
\multicolumn{1}{|l|}{BD-CSPN\cite{liu2020prototype}}                 & \textbf{59.99 $\pm$ 0.45} & \textbf{46.69} & \textbf{57.50} & \textbf{63.80} & \textbf{71.95} & \textbf{74.55 $\pm$ 0.37} \\
\multicolumn{1}{|l|}{TIM\cite{boudiaf2020transductive}}                     & \textbf{59.19 $\pm$ 0.43} & \textbf{46.74} & \textbf{56.76} & \textbf{63.03} & 70.21 & 73.09 $\pm$ 0.35 \\
\multicolumn{1}{|l|}{Transductive Finetuning\cite{dhillon2019baseline}} & 53.01 $\pm$ 0.40 & 41.98 & 51.74 & 56.31 & 62.82 & 65.27 $\pm$ 0.35 \\
\multicolumn{1}{|l|}{PT MAP\cite{hu2021leveraging}}                  & 56.74 $\pm$ 0.38 & 45.81 & 54.38 & 59.76 & 66.96 & 69.58 $\pm$ 0.37 \\ \hline
\end{tabular}
\caption{Top-1 accuracy of various few-shot learning methods on 1-shot tasks sampled from the tieredImageNet test set with uniform and semantic sampling strategies. For better-tieredImageNet, we show the average accuracy on the whole testbed, along with the average accuracy on the four quartiles of the testbed, when sorted by coarsity of the task (1st quartile contains the tasks with the smallest coarsity). For the testbed of uniformly sampled tasks, we only show the average accuracy on the whole testbed, as the results on each quartile are very similar to one another. We separate inductive (top) from transductive (bottom) methods. Best method(s) in each column is shown in bold.}
\label{tiered-1-shot}
\end{table*}

\begin{table*}[t!]

\centering
\begin{tabular}{c|ccccc|c|}
\cline{2-7}
                                              & \multicolumn{6}{c|}{5-shot}   \\ \cline{2-7}
                                              & \multicolumn{5}{c|}{better-tieredImageNet}                                                       & \multirow{2}{*}{Uniform testbed}    \\ \cline{2-6}
                                              & Whole testbed & 1st. Qtl           & 2nd. Qtl           & 3rd Qtl            & 4th Qtl            &     \\ \hline
\multicolumn{1}{|l|}{ProtoNet\cite{snell2017prototypical}}                & 70.77 $\pm$ 0.37 & 59.72 & 68.95 & 74.05 & 80.41 & 82.79 $\pm$ 0.25 \\
\multicolumn{1}{|l|}{Finetune\cite{Chen19}}                & 71.60 $\pm$ 0.37 & 60.63 & 69.68 & 74.86 & 81.26 & 83.66 $\pm$ 0.25 \\ \hline
\multicolumn{1}{|l|}{BD-CSPN\cite{liu2020prototype}}                  & 72.50 $\pm$ 0.37 & 61.23 & 70.47 & \textbf{75.94} & \textbf{82.38} & 84.70 $\pm$ 0.25 \\
\multicolumn{1}{|l|}{TIM\cite{boudiaf2020transductive}}                     & \textbf{73.32 $\pm$ 0.37} & \textbf{62.40} & \textbf{71.48} & \textbf{76.53} & \textbf{82.92} & \textbf{85.49 $\pm$ 0.24} \\
\multicolumn{1}{|l|}{Transductive Finetuning\cite{dhillon2019baseline}} & 70.79 $\pm$ 0.37 & 59.73 & 68.98 & 74.08 & 80.43 & 82.79 $\pm$ 0.25 \\
\multicolumn{1}{|l|}{PT MAP\cite{hu2021leveraging}}                  & 69.45 $\pm$ 0.36 & 58.97 & 67.40 & 72.36 & 79.09 & 81.54 $\pm$ 0.26 \\ \hline
\end{tabular}
\caption{Top-1 accuracy of various few-shot learning methods on 1-shot tasks sampled from the tieredImageNet test set with uniform and semantic sampling strategies. For better-tieredImageNet, we show the average accuracy on the whole testbed, along with the average accuracy on the four quartiles of the testbed, when sorted by coarsity of the task (1st quartile contains the tasks with the smallest coarsity). For the testbed of uniformly sampled tasks, we only show the average accuracy on the whole testbed, as the results on each quartile are very similar to one another. We separate inductive (top) from transductive (bottom) methods. Best method(s) in each column is shown in bold.}
\label{tiered-5-shot}
\end{table*}

\section{Experiments on new benchmarks}

\subsection{Implementation details}

We conducted the necessary experiments to bring out the need for novel few-shot classification benchmarks and showcase the limitations of state-of-the-art methods on more challenging settings. We restricted the comparison to methods allowing classical training (\ie non episodic) and to a unique set of hyper-parameters. This comparison will be enriched in future works. All parameters of our experiments can be found on our publicly available code \footnote{\url{https://github.com/sicara/semantic-task-sampling}}.

\paragraph{\textit{tieredImageNet}} We followed the original split of \cite{ren2018meta}. All methods tested in our benchmark use a common ResNet12 with parameters trained for 500 epochs with classical cross-entropy among the 351 classes of the train set using stochastic gradient descent with a batch size of 512 and learning rate of 0.1 with a decreasing factor of 0.1 after 350, 450 and 480 epochs. The trained weights are directly downloadable from our code. We built two testbeds with uniform class sampling (1-shot and 5-shot), and two testbeds (1-shot and 5-shot) with semantic task sampling (see Section \ref{semantic-task-sampling}) with $\alpha = 0.383$ and $\beta = 100.0$. These hyperparameters were selected to enforce the sampling of tasks with small coarsity while ensuring that all classes were equally represented in the testbed (with a small margin). This selection was monitored with visual observations shown in Figures \ref{fig:coarsity-better} and \ref{fig:occurrences}. We upsampled 10000 tasks, then we removed all duplicate tasks and downsampled them to 5000 tasks. All tasks present 10 queries per class. 

\paragraph{Danish Fungi 2020} All methods tested in our benchmark use the built-in ResNet18 from \texttt{PyTorch} with weights trained on ImageNet. Since DF20 is already fine-grained, we built four 5000-tasks testbeds (5-way 1-shot, 5-way 5-shot, 100-way 1-shot, and 100-way 5-shot) with uniform class sampling. All tasks present 10 queries per class.

\paragraph{Methods} For Finetune \cite{Chen19} we use 10 fine-tuning steps with a learning rate of $10^{-3}$. For Transductive Information Maximization (TIM) \cite{boudiaf2020transductive} we use 100 fine-tuning steps with a learning rate of $10^{-3}$ and put a $0.1$ weight on the conditional entropy term of the loss. For Transductive Finetuning \cite{dhillon2019baseline} we use 25 inference steps with a learning rate of $5 \times 10^{-5}$. These hyperparameters were selected to fit the original implementations of these methods. We insist that we didn't put any additional effort into further optimizing any few-shot method on our benchmarks.

\begin{table*}[ht]
\centering
\begin{tabular}{l|cc|cccc|}
\cline{2-7}
                                              & \multicolumn{2}{c|}{5-way}                               & \multicolumn{4}{c|}{100-way}                                     \\ \cline{2-7} 
                                              & \multicolumn{1}{c|}{1-shot}           & 5-shot           & \multicolumn{2}{c|}{1-shot}        & \multicolumn{2}{c|}{5-shot} \\ \cline{2-7} 
                                              & \multicolumn{2}{c|}{Top-1}                               & Top-1 & \multicolumn{1}{c|}{Top-5} & Top-1        & Top-5        \\ \hline
\multicolumn{1}{|l|}{ProtoNet\cite{snell2017prototypical}}  & \multicolumn{1}{c|}{37.55 $\pm$ 0.25} & 60.53 $\pm$ 0.27 &7.81 $\pm$ 0.08 & \multicolumn{1}{c|}{20.12 $\pm$ 0.13}& 17.69 $\pm$ 0.12 &  40.20 $\pm$ 0.16 \\
\multicolumn{1}{|l|}{Finetune\cite{Chen19}}                & \multicolumn{1}{c|}{47.00 $\pm$ 0.30} & 65.06 $\pm$ 0.29 & \textbf{9.70 $\pm$ 0.09} & \multicolumn{1}{c|}{25.94 $\pm$ 0.15} & \textbf{19.60 $\pm$ 0.12} & \textbf{44.25 $\pm$ 0.17} \\ \hline
\multicolumn{1}{|l|}{BD-CSPN\cite{liu2020prototype}}   & \multicolumn{1}{c|}{47.81 $\pm$ 0.33} & \textbf{66.32 $\pm$ 0.30} & \textbf{9.75 $\pm$ 0.09} & \multicolumn{1}{c|}{24.11 $\pm$ 0.15} & \textbf{19.52 $\pm$ 0.13} & 41.79 $\pm$ 0.17 \\
\multicolumn{1}{|l|}{TIM\cite{boudiaf2020transductive}}    & \multicolumn{1}{c|}{40.73 $\pm$ 0.28} & 62.89 $\pm$ 0.28 & 8.36 $\pm$ 0.09 & \multicolumn{1}{c|}{21.30 $\pm$ 0.14} & 18.53 $\pm$ 0.12 & 41.47 $\pm$ 0.17 \\
\multicolumn{1}{|l|}{Trans. Finetuning\cite{dhillon2019baseline}} & \multicolumn{1}{c|}{37.54 $\pm$ 0.25} & 60.54 $\pm$ 0.27 & 7.71 $\pm$ 0.08 & \multicolumn{1}{c|}{20.13 $\pm$ 0.13} & 17.69 $\pm$ 0.12 & 40.21 $\pm$ 0.16 \\
\multicolumn{1}{|l|}{PT MAP\cite{hu2021leveraging}}   & \multicolumn{1}{c|}{\textbf{52.08 $\pm$ 0.35}} & \textbf{66.78 $\pm$ 0.29} & 9.54 $\pm$ 0.09 & \multicolumn{1}{c|}{\textbf{26.37 $\pm$ 0.15}}  & 18.50 $\pm$ 0.12 & 43.36 $\pm$ 0.16 \\ \hline
\end{tabular}
\caption{Accuracy of various few-shot learning methods on DF20. For 100-way tasks, we report both top-1 and top-5 accuracy. For 5-way tasks, we do not report top-5 accuracy as we found that it was always 100\%. Best method(s) in each column is shown in bold.}
\label{results-fungi}
\end{table*}

\subsection{Results}
\label{sec:results}

\paragraph{\textit{tieredImageNet}} Results for \textit{tiered}ImageNet are shown in Tables \ref{tiered-1-shot} and \ref{tiered-5-shot}. The immediate observation that we can make is that our benchmark \textit{better-tiered}ImageNet is much more challenging than uniform task sampling, with a performance drop of 12 to 15\% in top-1 accuracy for all settings and methods. For further details, we sorted all 5000 tasks with respect to their coarsity and grouped them into four quartiles. The 1st quartile contains the most fine-grained tasks, and the 4th quartile contains the coarsest tasks. From these results, we can confirm that coarsity is indeed correlated to the difficulty of the task since the performance consistently improves when moving towards coarser tasks. We finally observed that even the 4th quartile seems to be more challenging than the uniform benchmark. This is consistent with the demography of tasks shown in Figure \ref{fig:coarsities}, since the tasks constituting the 4th quartile of our testbed show a smaller average coarsity than the uniform testbed.

We observed that transductive methods\cite{liu2020prototype, boudiaf2020transductive, dhillon2019baseline, hu2021leveraging}, which use the unlabeled information from the query set, unsurprisingly show the best results on both set-ups but especially in 1-shot classification. The leaderboard seems to be consistent on all quartiles, suggesting that none of these methods are "specialized" towards a particular demographic of tasks.

\paragraph{Danish Fungi 2020} Results for DF20 are shown in Table \ref{results-fungi}. They show that while being more challenging than \textit{tiered}ImageNet and \textit{better-tiered}ImageNet, our DF20 benchmark still constitutes an achievable task. We also report results showing that all methods struggle in the more challenging problem of 100-way classification, especially in the 1-shot setting (less than 10\% top-1 accuracy for the best model, less than 20\% in the 5-shot setting). We believe that this should stand as a red flag regarding the ability of state-of-the-art few-shot classification methods to scale to real-life use cases.

\begin{figure}
    \centering
    \includegraphics[width=\linewidth]{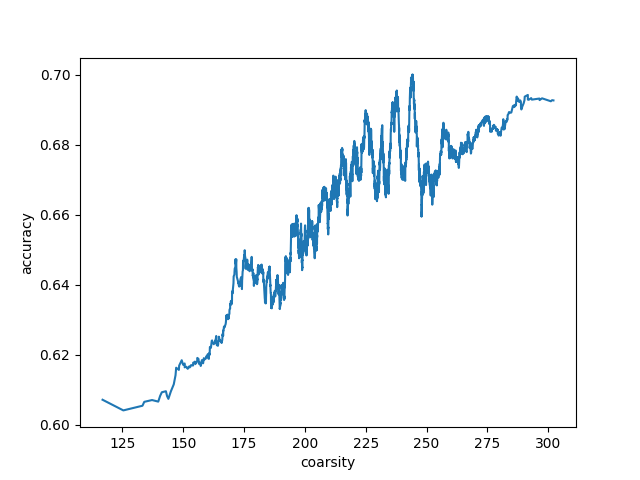}
    \caption{Correlation between the coarsity of the 5000 tasks sampled from DF20 for our 5-way 5-shot testbed and the accuracy of the PT-MAP\cite{hu2021leveraging} method. The plot is smoothed using the rolling average over a window of 200 tasks.}
    \label{fig:fungi-acc-coarsity}
\end{figure}

To complete the study, we show in Figure \ref{fig:fungi-acc-coarsity} the correlation between the coarsity of a task (based on the taxonomy of DF20) and the accuracy of the PT-MAP\cite{hu2021leveraging} method. We observe that, as was the case for \textit{better-tiered}ImageNet, the closest the classes sampled from Fungi are from one another, the harder the task composed of these classes.

\section{Conclusion}

We showed that the widely used \textit{tiered}ImageNet benchmark with uniform sampling of classes led to evaluating few-shot learning models on disproportionately coarse tasks. We used semantic task sampling to generate a more informative testbed from \textit{tiered}ImageNet's test set. We also pushed forward as a new benchmark for Few-shot Image Classification models the Danish Fungi 2020 dataset, which we believe to be an incredibly promising playing field for future research in Few-Shot Learning. Finally, we showed that state-of-the-art methods dramatically fail when confronted with many-way classification (here many is 100) of fine-grained objects. We insist that this setting is \textbf{not} far-fetched and fits tangible industrial use cases. We believe that these results should push us to take a step back and re-assess the way we currently think about Few-Shot Image Classification.

We used to define a few-shot classification task by its number of ways and its number of shots, addressing $n$-way $k$-shot classification as an indivisible problem. What we did here can be seen as a novel framework, in which the number of classes is not sufficient to define a task: we need to know what these classes are. In the same fashion, future works may go beyond tasks defined by their number of shots, and consider which images are chosen for the support set.

In this work, we addressed what we believe to be a very limiting bias of current Few-Shot Learning benchmarks \ie a bias towards coarse tasks. We chose to tackle this particular shortcoming because we observed that it was the main difference between academic benchmarks and the industrial applications of Few-Shot Learning that we encountered. However, many more limitations of few-shot learning benchmarks are yet to address: the fixed shape of the tasks, the strict balance in both support and query sets, the empty overlap between large-scale classes (currently only used for base training) and few-shot classes, no prior in the choice of support instances, and many other of which we did not think yet. We believe that addressing these shortcomings must be considered a priority in our field, and we encourage any and all who agree to join us in this effort.

{\small
\bibliographystyle{ieee_fullname}
\bibliography{egbib}
}

\end{document}